\documentclass[10pt,twocolumn,letterpaper]{article}
\usepackage[pagenumbers]{cvpr}
\usepackage{iftex}
\ifPDFTeX
  \usepackage[T1]{fontenc}
\else
  \usepackage[no-math]{fontspec}
\fi
\usepackage{amsmath,amssymb,mathtools,bm}
\usepackage{booktabs,multirow,array}
\usepackage{xspace}
\usepackage[breaklinks,hidelinks]{hyperref}
\AtBeginDocument{%
  \everymath\expandafter{\the\everymath
    \medmuskip=4mu\relax\thickmuskip=4mu\relax}%
  \frenchspacing}
\newcommand{\method}{TAP\xspace}
\newcommand{\Aavg}{\overline{A}}
\newcommand{\AT}{A_T}
\newcommand{\R}{\mathbb{R}}
\DeclareMathOperator*{\argminop}{arg\,min}
\DeclareMathOperator*{\argmaxop}{arg\,max}
\title{TAP: Tapping the Discriminative Potential of\\
Frozen Representations for Class-Incremental Learning}
\author{{\fontsize{11}{13}\selectfont Zihao Wang$^{1}$\enspace Yu-Wei Zhan$^{2}$\enspace Xiao-Qian Liu$^{1}$\enspace Zhen-Duo Chen$^{1}$\enspace Weitong Liu$^{1}$\enspace Xin Luo$^{1,*}$\enspace Xin-Shun Xu$^{1}$}\\
{\small $^{1}$Shandong University, Jinan, Shandong, China}\\
{\small $^{2}$Zhejiang University of Technology, Hangzhou, Zhejiang, China}\\
{\small $^{*}$Corresponding author: \href{mailto:luoxin.lxin@gmail.com}{luoxin.lxin@gmail.com}}}
\begin{document}
\setlength{\dbltextfloatsep}{13pt plus 2pt minus 2pt}
\setlength{\textfloatsep}{13pt plus 2pt minus 2pt}
\maketitle
\begin{abstract}
Pre-trained models provide transferable representations
for class-incremental learning (CIL). However, sequentially adapting these
representations introduces optimization costs and may disrupt old-class
feature structure. This motivates us to explore whether the discriminative potential
of these representations can be further exploited by continually accumulating class information
while keeping the encoders frozen. To this end, we propose \method, a new and novel framework for tapping the discriminative
potential of frozen representations in class-incremental learning.
It accumulates visual and semantic class information in the frozen CLIP
embedding space with only lightweight optimization.
\method comprises three modules: class-conditional visual distribution modeling,
class-adaptive semantic calibration, and visual-semantic posterior fusion.
The first module retains class means and within-class scatters,
combining class-specific and shared covariances to stabilize distribution
estimation. The second module learns
class-specific weights over text templates from current-task images,
adapting text representations to downstream classes.
The third module combines
the visual and semantic posteriors through a weighted product of experts
for prediction over all seen classes without retaining images from previous tasks.
Across nine datasets under two incremental
protocols, \method achieves state-of-the-art performance, attaining the
highest average incremental and final-stage accuracies among the compared
methods.
The code is provided as an ancillary file.
\end{abstract}

\section{Introduction}
\label{sec:introduction}

Class-incremental learning (CIL) seeks to learn new classes sequentially
while retaining recognition of prior ones. The exemplar-free protocol has
gained prominence in CIL, as it prohibits storing any old-class training
exemplars for rehearsal and requires the model to maintain old-class
recognition performance after all old-class images are discarded
post-training. Large-scale pre-trained models offer transferable
representations for this challenging setup and have emerged as a major
technical route.

Class-incremental learning with pre-trained models explores both how to
adapt pre-trained representations to downstream tasks and how to directly
exploit their existing discriminative ability. The former introduces task
information through prompt learning~\cite{wang2022l2p,smith2023coda} or
adapters~\cite{zhou2024ease,luo2025lada}. However, continual updates to shared parameters
may degrade old-class representations, and storing task-specific modules
incurs extra storage and computational overheads. The latter constructs class
prototypes from frozen features~\cite{zhou2025revisiting}, but representing
each class solely by its mean feature fails to explicitly capture within-class
variation. This motivates us to ask: \emph{can we improve incremental
recognition by more fully accumulating and exploiting visual and semantic
class information while keeping the encoders frozen?}

A frozen feature space enables the preservation of class information via
prototypes~\cite{zhou2025revisiting} or class-specific
covariances~\cite{goswami2023fecam}. Covariances encode within-class
variation beyond class centers, yet their estimation tends to be unreliable
under limited training samples~\cite{ledoit2004covariance}, which calls for more robust estimation of class distributions.
Beyond storing visual information via class distributions, CLIP offers
semantic cues for classification by aligning images with class text
representations. However, the effectiveness of different text templates
varies across classes. This motivates us to explore
class-adaptive combinations of text templates.

To this end, we propose \method, a framework that unlocks the discriminative
potential of frozen representations for class-incremental learning.
It keeps the CLIP encoders frozen and requires only lightweight optimization.
The framework comprises three modules: class-conditional visual distribution
modeling, class-adaptive semantic calibration, and visual-semantic posterior
fusion. The visual module builds the visual expert by merging
class-specific and shared within-class covariances via
shrinkage~\cite{friedman1989rda}, and then computes Gaussian discriminant scores
that account for both distance and distribution volume. Thanks to the stored class statistics, this visual expert can be updated
without access to old images or individual feature vectors. The semantic module builds
the semantic expert by weighting a shared pool of text templates for each
class. Current-task images supervise new-class template weights.
The fusion module combines the two experts'
posteriors through a weighted product of experts~\cite{hinton2002poe}.
Both encoders remain frozen; only the template weights associated with new classes
require gradient-based optimization.

Our main contributions are as follows:
\begin{itemize}
    \item We propose \method, which continually accumulates class information
    via class statistics and calibrated text representations in a frozen
    CLIP embedding space. This design enables incremental recognition without
    encoder adaptation or old-image replay.
    \item We leverage shared within-class statistics to stabilize class-specific
    covariance estimates and adapt text representations for downstream classes
    via template weighting supervised by current-task images. We further
    fuse visual and semantic posteriors to combine the two complementary sources of
    evidence for prediction.
    \item We achieve state-of-the-art performance across nine datasets under
    two incremental protocols, attaining the highest cross-dataset average
    incremental and final-stage accuracies among all compared methods.
\end{itemize}

\section{Related Work}
\label{sec:related}

\subsection{Pre-trained Models for CIL}
Large-scale pre-trained models provide transferable representations that can
serve as a foundation for downstream learning. Accordingly,
class-incremental learning methods built upon pre-trained models seek to
leverage these representations to learn new classes while preserving recognition performance
on previously seen classes~\cite{zhou2024ptmsurvey}.
For pre-trained visual models, such as vision transformers (ViTs)~\cite{dosovitskiy2021vit}, incremental
adaptation can operate through auxiliary modules or updates to the backbone~\cite{gao2023lae,he2025cllora,wang2025tuna,zhang2026mobo}.
With the backbone frozen,
L2P~\cite{wang2022l2p}, DualPrompt~\cite{wang2022dualprompt}, and
CODA-Prompt~\cite{smith2023coda} learn prompts, while
EASE~\cite{zhou2024ease} learns task-specific adapters and aggregates their
feature subspaces.
For backbone adaptation,
SLCA~\cite{zhang2023slca} applies a lower learning rate for the representation
layers than for the classifier to
preserve the generalizability of the pre-trained features.

While the above methods operate solely within visual representation space,
vision-language models such as CLIP~\cite{radford2021clip} further align
visual representations with textual descriptions, which allows recognizing
new classes from their text embeddings without encoder updates~\cite{thengane2022continualclip}.
Beyond zero-shot recognition, textual semantics can guide the adaptation of
visual representations to incremental tasks~\cite{zhou2025proof,huang2024rapf,huang2025mgclip}.
Textual descriptions also provide complementary semantic information for
class discrimination~\cite{li2026bofa,zhou2025engine,yu2025clgcbm,xie2026area}.
Parameter-efficient continual updates can also be constrained through gradient
projection~\cite{qiao2025pegp}, with task-aware prompt projection explored in
federated CIL~\cite{ke2025tppr}.
Most of these methods introduce task-specific representation adaptation,
adding optimization cost during CIL. In contrast, we show that competitive incremental
recognition can be achieved without such adaptation, by combining class-wise
visual statistics with calibrated class text representations from frozen CLIP
encoders through visual-semantic posterior fusion.

\subsection{Prototype and Distribution-Based CIL}
Beyond representation adaptation, large-scale pre-trained models can also
provide discriminative features for downstream classes through their
generalization ability. These features provide a basis for constructing
class prototypes.
Using the mean feature of each class as its prototype supports incremental
recognition without old-image replay~\cite{zhou2025revisiting}.
To also account for feature correlations in classification, accumulated
class-wise feature sums can be combined with a feature Gram matrix to obtain
closed-form classifier weights in a randomly projected and nonlinearly
transformed feature space~\cite{mcdonnell2023ranpac}.

The mean-based class prototypes described above represent class centers in feature space,
while class-conditional distributions additionally characterize how samples
scatter around these centers~\cite{hayes2020slda}. Class-specific covariance
matrices capture feature variances and correlations, allowing distance measures to
reflect each class's pattern of feature variation~\cite{goswami2023fecam,khawand2023srda},
while Gaussian mixtures use multiple component means, covariances, and
mixing weights to describe more complex within-class structure~\cite{korycki2024mix,cao2026statistics}.

Despite these richer distribution modeling capabilities, statistical
classification within high-dimensional feature spaces confronts a fundamental
trade-off: capturing class-specific variation versus reliably estimating
comparable class scores from limited samples.
Our framework shrinks class-specific covariance estimates toward a shared
covariance and accounts for both distance and distribution
volume when computing class scores. This design stabilizes estimation and
improves score comparability under limited training samples.

\begin{figure*}[t]
\centering
\includegraphics[width=\textwidth]{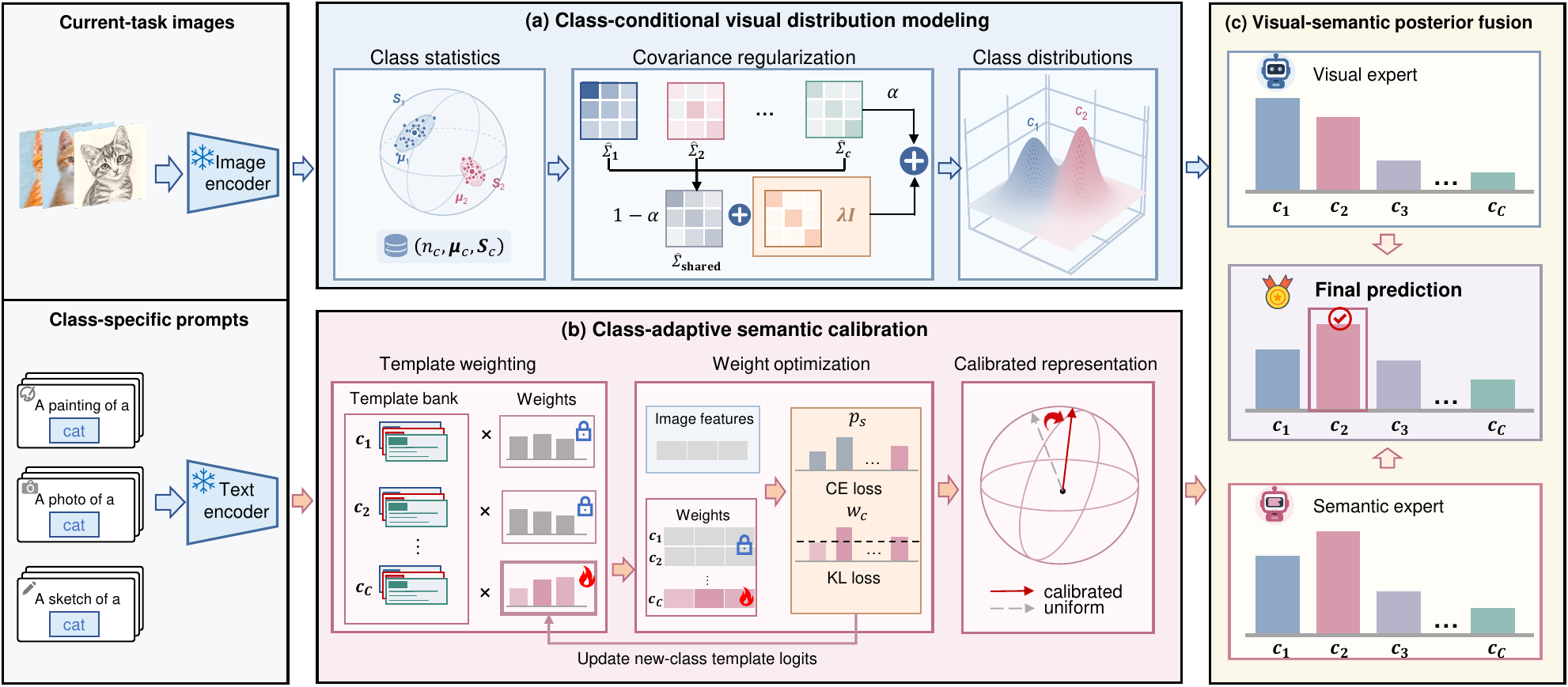}
\caption{Overview of \method: (a) visual distribution modeling estimates
class means and regularized covariances; (b) semantic calibration learns
class-specific template weights; and (c) posterior fusion combines visual
and semantic predictions.
Both CLIP encoders remain frozen. Surfaces illustrate class densities.}
\label{fig:method_framework}
\end{figure*}

\section{Preliminaries}
\label{sec:preliminaries}

\subsection{Problem Definition of CIL}
Class-incremental learning considers a sequence of $T$ tasks. At stage
$t\in\{1,\ldots,T\}$, the model receives a labeled training set
$\mathcal{D}_t=\{(\bm{x}_i,y_i)\}_{i=1}^{N_t}$, where $N_t$ is the number of
training samples and
$\bm{x}_i$ and $y_i$ are the $i$-th training image and its class label.
Let $\mathcal{C}_t^{\mathrm{new}}$ denote the new classes introduced at
stage $t$, so that $y_i\in\mathcal{C}_t^{\mathrm{new}}$.
Class sets from different tasks are disjoint. The set of all seen classes is
$\mathcal{C}_t=\mathcal{C}_1^{\mathrm{new}}\cup\cdots\cup
\mathcal{C}_t^{\mathrm{new}}$.
We denote the number of seen classes at stage $t$ by $\lvert\mathcal{C}_t\rvert$.
We follow the exemplar-free setting: only $\mathcal{D}_t$ is available for
training at stage $t$, and images from earlier tasks cannot be stored or replayed.
Let $\mathcal{P}_t$ denote the joint distribution of image-label pairs
$(\bm{x},y)$ over all seen classes $\mathcal{C}_t$ at stage $t$.
The objective at this stage is to learn a classifier $h_t$ that minimizes
the expected classification error under $\mathcal{P}_t$:
\begin{equation}
    h_t^*=\argminop_{h_t}
    \mathbb{E}_{(\bm{x},y)\sim\mathcal{P}_t}
    \left[\mathbb{I}\bigl(h_t(\bm{x})\ne y\bigr)\right],
    \label{eq:cil_objective}
\end{equation}
where $h_t^*$ denotes an optimal classifier under $\mathcal{P}_t$,
$\mathbb{E}$ denotes expectation over $\mathcal{P}_t$, and
$\mathbb{I}(\cdot)$ is the indicator function, equal to one when its
condition is true and zero otherwise. The classifier $h_t$ predicts a
label in $\mathcal{C}_t$ without access to the image's task identity.
Thus, the goal is to learn new classes while maintaining recognition
performance on previously seen classes.

\subsection{CLIP-Based Class-Incremental Learning}
In CLIP-based class-incremental learning, images are classified over an
expanding set of seen classes. CLIP~\cite{radford2021clip} aligns image and
text representations in a shared $d$-dimensional feature space, allowing
class text representations to serve as classification references as new
classes arrive.
Its image encoder produces an $\ell_2$-normalized
embedding $\bm{z}_{\mathrm{img}}\in\R^d$ for an input image $\bm{x}$.
Its text encoder produces an $\ell_2$-normalized embedding
$\bm{z}_{\mathrm{text}}^{c}\in\R^d$ from a description of class $c$.
In this shared space, the cosine similarity
$\bm{z}_{\mathrm{img}}^\top \bm{z}_{\mathrm{text}}^{c}$ quantifies the
compatibility of image $\bm{x}$ with class $c$, where $\top$ denotes transpose.
Scaling these similarities by the inverse of a positive temperature $\tau$
and applying
softmax over all seen classes yields the class probability:
\begin{equation}
    p_{\mathrm{CLIP}}(c\mid \bm{z}_{\mathrm{img}})
    =\frac{\exp(\bm{z}_{\mathrm{img}}^\top \bm{z}_{\mathrm{text}}^{c}/\tau)}
    {\sum_{j\in\mathcal{C}_t}\exp(\bm{z}_{\mathrm{img}}^\top \bm{z}_{\mathrm{text}}^{j}/\tau)}.
    \label{eq:clip_classification}
\end{equation}
The predicted label is $\hat y=\argmaxop\nolimits_{c\in\mathcal{C}_t}
p_{\mathrm{CLIP}}(c\mid \bm{z}_{\mathrm{img}})$, corresponding to the class with
the highest probability.

At each incremental stage, labeled training images of the new classes can
also be used to update a visual classifier by learning classification
weights, constructing class prototypes, or estimating class-conditional
distributions.
Let $s_c^v(\bm{z}_{\mathrm{img}})$ denote the visual classification score for
class $c$. With a positive temperature $\tau_v$, these scores can likewise
be normalized across all seen classes:
\begin{equation}
    p_v(c\mid \bm{z}_{\mathrm{img}})
    =\frac{\exp(s_c^v(\bm{z}_{\mathrm{img}})/\tau_v)}
    {\sum_{j\in\mathcal{C}_t}\exp(s_j^v(\bm{z}_{\mathrm{img}})/\tau_v)}.
    \label{eq:visual_classification}
\end{equation}
Here $\tau_v>0$ controls the concentration of the visual predictive
distribution: smaller values concentrate probability on higher-scoring
classes, while larger values produce a flatter distribution.
The predicted label is likewise $\hat y=\argmaxop\nolimits_{c\in\mathcal{C}_t}
p_v(c\mid \bm{z}_{\mathrm{img}})$.
As new classes arrive, their training images are used to update the visual
classifier, which can deliver predictions without directly relying on text
embeddings.

% Keep the method equations together before the experiments page.
\begingroup
\addtolength{\abovedisplayskip}{-2pt}
\addtolength{\belowdisplayskip}{-2pt}
\addtolength{\abovedisplayshortskip}{-1pt}
\addtolength{\belowdisplayshortskip}{-1pt}
\section{Method}
\label{sec:method}

To further exploit the discriminative potential of pre-trained
representations, we propose \method, which supports incremental recognition
by continually accumulating class information in a frozen CLIP embedding space.
Keeping both encoders frozen makes class statistics collected at different
stages comparable in the same embedding space and preserves the pre-trained
image-text alignment. Class information can therefore be accumulated
through visual statistics and calibrated text representations without
changing the feature coordinates used by previous tasks.

As illustrated in Figure~\ref{fig:method_framework}, \method comprises
three interconnected modules: class-conditional visual distribution modeling,
class-adaptive semantic calibration, and visual-semantic posterior fusion.
The visual module builds the visual expert by modeling each class's feature
center and within-class variation,
using shared covariance to stabilize statistical estimation.
The semantic module builds the semantic expert by
learning template weights from current-task images to adapt pre-trained
text representations to downstream classes.
The fusion module combines the classification
evidence from the two experts, where the visual expert provides the visual
posterior $p_v$ and the semantic expert delivers the calibrated semantic
posterior $p_s$, to produce predictions, exploiting the complementary
information encoded in visual distributions and textual semantics.
The framework requires no old-image replay and applies lightweight
optimization only to template weights.

\subsection{Class-Conditional Visual Distribution Modeling}
\label{sec:visual_model}
To account for both class centers and within-class variation in visual
classification, we estimate class-conditional feature distributions in the
frozen CLIP embedding space.
The class mean locates the center of a class's image features in the
embedding space, while the within-class scatter describes the magnitude of
their variation around this center along different directions. Together,
these statistics provide the basis for the visual decision rule.
For each new class $c$ with $n_c$ training images, let
$\bm{z}_{\mathrm{img}}^{c,i}$ denote the normalized feature of its $i$-th image.
The class mean and within-class scatter are:
\begin{equation}
\begin{aligned}
\bm{\mu}_c&=\frac{1}{n_c}\sum_{i=1}^{n_c}\bm{z}_{\mathrm{img}}^{c,i},\\
\bm{S}_c&=\sum_{i=1}^{n_c}
(\bm{z}_{\mathrm{img}}^{c,i}-\bm{\mu}_c)(\bm{z}_{\mathrm{img}}^{c,i}-\bm{\mu}_c)^\top,
\end{aligned}
\label{eq:visual_statistics}
\end{equation}
where $\bm{\mu}_c\in\R^d$ is obtained by averaging the image features of class
$c$, and the within-class scatter $\bm{S}_c\in\R^{d\times d}$ is computed by
summing the outer products of their deviations from $\bm{\mu}_c$.
After each task, the retained class information consists of
$(n_c,\bm{\mu}_c,\bm{S}_c)$; images and individual embeddings are discarded.

The class-specific covariance is estimated from that class's own training
features to characterize the directions and magnitudes of variation around
$\bm{\mu}_c$. However, this estimate is susceptible to sampling
noise when the number of training samples is small relative to the feature
dimension. Since different classes may exhibit partly similar patterns of
within-class variation, aggregating their within-class scatters yields a shared
covariance that serves as a more stable reference for estimation.
Shrinking class-specific covariances toward
this reference suppresses fluctuations induced by limited
training samples while retaining class-specific variation. The class-specific
and shared covariances are computed from the individual and aggregated scatters,
respectively:
\begin{equation}
\widehat{\bm{\Sigma}}_c=\frac{\bm{S}_c}{n_c-1},\qquad
\widehat{\bm{\Sigma}}_{\mathrm{shared}}
=\frac{\sum_{c\in\mathcal C_t}\bm{S}_c}
{\sum_{c\in\mathcal C_t}(n_c-1)}.
\label{eq:pooled_covariance}
\end{equation}
The class-specific covariance is obtained by dividing its scatter by $n_c-1$.
The shared covariance is computed by summing the scatters of all seen classes
and dividing by their total within-class degrees of freedom,
$\sum_{c\in\mathcal C_t}(n_c-1)$. It is therefore a
weighted average of the class-specific covariances, with weights proportional to
$n_c-1$.
At each new task, the shared covariance is updated by adding the new classes'
counts and scatters to the retained totals, while old-class statistics remain
unchanged.

Each empirical class-specific covariance is combined with the shared
covariance~\cite{friedman1989rda}, and a positive diagonal term is added
to ensure invertibility:
\begin{equation}
\bm{\Sigma}_c^{\mathrm{mix}}
=(1-\alpha)\widehat{\bm{\Sigma}}_{\mathrm{shared}}
+\alpha\widehat{\bm{\Sigma}}_c+\lambda\bm{I}.
\label{eq:mixed_covariance}
\end{equation}
Here the covariance mixing coefficient $\alpha\in[0,1]$ controls the relative contributions of class-specific
and shared covariances, $\lambda>0$ controls diagonal regularization, and
$\bm{I}\in\R^{d\times d}$ is the identity matrix.
The regularized covariance $\bm{\Sigma}_c^{\mathrm{mix}}$ in Eq.~\eqref{eq:mixed_covariance} is then recomputed for every seen class,
incorporating the updated shared covariance into old classes' regularized
covariances without accessing old images.

For classification, each class's feature distribution is approximated
by a Gaussian with mean $\bm{\mu}_c$ and regularized
covariance $\bm{\Sigma}_c^{\mathrm{mix}}$. Assuming equal prior probabilities
for all seen classes and omitting constants common to all classes, the
Gaussian discriminant score is:
\begin{equation}
\begin{aligned}
s_c^v(\bm{z}_{\mathrm{img}})
&=-\frac12(\bm{z}_{\mathrm{img}}-\bm{\mu}_c)^\top
(\bm{\Sigma}_c^{\mathrm{mix}})^{-1}(\bm{z}_{\mathrm{img}}-\bm{\mu}_c)\\
&\quad-\frac12\log|\bm{\Sigma}_c^{\mathrm{mix}}|,
\end{aligned}
\label{eq:visual_discriminant}
\end{equation}
where $|\cdot|$ denotes the determinant.
The first (quadratic) term measures the squared Mahalanobis distance between the
image feature and the class mean, assigning smaller penalties to deviations
along high-variance directions; the second (log-determinant) term accounts for
distribution volume, counteracting the scoring advantage that broader
distributions would otherwise gain from smaller distances.
Substituting these scores into Eq.~\eqref{eq:visual_classification} normalizes
them over all seen classes to obtain $p_v(c\mid \bm{z}_{\mathrm{img}})$, the
probability assigned to class $c$ by the visual expert for image embedding
$\bm{z}_{\mathrm{img}}$.

\subsection{Class-Adaptive Semantic Calibration}
\label{sec:semantic_calibration}
CLIP classification commonly relies on predefined text templates to convert class
names into textual descriptions. We use publicly available templates and
dataset-specific variants, yet their compatibility with downstream images
may differ across classes. To adapt class text
representations to these images, we learn class-specific template weights
using current-task training images. All classes
within a dataset share a fixed pool of $K$ templates, while each class learns
its own combination weights. The template embeddings and both
encoders remain frozen.

For class $c$, let $\bm{z}_{\mathrm{text}}^{c,k}\in\R^d$ denote the normalized text
embedding obtained by inserting its name into template $k$, with
$k\in\{1,\ldots,K\}$. Learnable template logits $\bm{a}_c\in\R^K$ are
initialized at zero, corresponding to uniform weights. These logits
determine the template weights through a softmax, and the resulting weights
are used to construct the class text representation:
\begin{equation}
\begin{aligned}
&\bm{w}_c=\operatorname{softmax}(\bm{a}_c),\\
&\widetilde{\bm{z}}_{\mathrm{text}}^{c}
=\operatorname{norm}\left(\sum_{k=1}^{K}w_{c,k}\bm{z}_{\mathrm{text}}^{c,k}\right),
\end{aligned}
\label{eq:template_calibration}
\end{equation}
where $\bm{w}_c\in\R^K$ is the learned template-weight distribution, with
positive entries summing to one.
The template embeddings are summed with weights $w_{c,k}$, and the resulting
vector is normalized to unit length to obtain the calibrated class text representation
$\widetilde{\bm{z}}_{\mathrm{text}}^{c}$. Here $\operatorname{norm}(\cdot)$ denotes
dividing a vector by its $\ell_2$ norm.

The calibrated text representation $\widetilde{\bm{z}}_{\mathrm{text}}^{c}$ is used to compute the image-text
matching score
$s_c^s(\bm{z}_{\mathrm{img}})=\bm{z}_{\mathrm{img}}^\top\widetilde{\bm{z}}_{\mathrm{text}}^{c}$.
Dividing these scores by a semantic temperature $\tau>0$ and applying
softmax over all seen classes, as in Eq.~\eqref{eq:clip_classification},
yields the semantic posterior $p_s(c\mid \bm{z}_{\mathrm{img}})$.
The temperature $\tau$ controls the distribution's concentration.
To better align the weighted text representations with current-task images,
we supervise these probabilities using ground-truth class labels to
learn the new classes' template logits. The objective combines a
classification loss with a template-weight regularizer:
\begin{equation}
\begin{aligned}
\mathcal L_{\mathrm{sem}}
&=-\frac1{N_t}\sum_{i=1}^{N_t}
\log p_s(y_i\mid \bm{z}_{\mathrm{img}}^{i})\\
&\quad+\frac{\beta}{|\mathcal C_t^{\mathrm{new}}|}
\sum_{c\in\mathcal C_t^{\mathrm{new}}}
\operatorname{KL}\left(\bm{w}_c\,\Vert\,\bm 1/K\right).
\end{aligned}
\label{eq:template_loss}
\end{equation}
The classification loss encourages a higher semantic probability for the
true class of each training image, whose normalized feature is denoted
by $\bm{z}_{\mathrm{img}}^{i}$, thereby adapting the template weights to
current-task images. To prevent excessive concentration on a few templates,
the regularizer penalizes deviations from the uniform distribution
$\bm 1/K$ using Kullback--Leibler (KL) divergence.
Here $\bm 1\in\R^K$ is the all-ones vector, and $\beta\geq0$ controls the
regularization strength.

Under this objective, only the template logits of newly introduced classes
are optimized at each task; text representations of old classes remain
unchanged and both CLIP encoders stay frozen. After optimization, the weighted text representation and its $K$
template weights are retained for each new class. The text representations
of new and old classes together form the semantic expert, whose
predicted probabilities are subsequently fused with those of the visual
expert.

\subsection{Visual-Semantic Posterior Fusion}
\label{sec:posterior_fusion}
The fusion module combines predictions from two complementary experts,
producing the visual posterior $p_v$ and the calibrated semantic posterior
$p_s$, respectively. The visual expert characterizes downstream image distributions via
class means and within-class variation, whereas the semantic expert uses
calibrated text representations and pre-trained image-text associations
for classification. These different sources of evidence often produce
different class rankings for the same image. We therefore combine their
predictive distributions over all seen classes, allowing the final decision
to account for both visual distribution structure and textual semantics.
Specifically, we adopt a weighted product of experts (PoE)~\cite{hinton2002poe}
with adjustable relative contributions from the two experts:
\begin{equation}
q_\gamma(c\mid \bm{z}_{\mathrm{img}})=
\frac{p_v(c\mid \bm{z}_{\mathrm{img}})^{1-\gamma}
      p_s(c\mid \bm{z}_{\mathrm{img}})^\gamma}
{\sum_{j\in\mathcal C_t}p_v(j\mid \bm{z}_{\mathrm{img}})^{1-\gamma}
      p_s(j\mid \bm{z}_{\mathrm{img}})^\gamma}.
\label{eq:posterior_fusion}
\end{equation}
Here $q_\gamma(c\mid \bm{z}_{\mathrm{img}})$ is the fused probability of class $c$
for image embedding $\bm{z}_{\mathrm{img}}$, and the fusion coefficient $\gamma\in[0,1]$ controls the
relative weights of the two experts. After taking the logarithm of the
fused probability and omitting the normalization term shared by all
classes, the final prediction can be written as:
\begin{equation}
\begin{aligned}
\hat y=\argmaxop\nolimits_{c\in\mathcal C_t}\bigl[
&(1-\gamma)\log p_v(c\mid \bm{z}_{\mathrm{img}})\\
&+\gamma\log p_s(c\mid \bm{z}_{\mathrm{img}})\bigr].
\end{aligned}
\label{eq:poe_prediction}
\end{equation}

Weighted product fusion can also be understood as solving an optimization
problem: for a given image, find a probability distribution $q$ over all
seen classes that minimizes the weighted sum of KL divergences to the
visual and semantic experts' predictive distributions. The fused
distribution $q_\gamma$ defined above is the unique solution to this problem:
\begin{equation}
q_\gamma=\argminop_{\substack{q(c)\geq0\\\sum_{c\in\mathcal C_t}q(c)=1}}
\left[(1-\gamma)\operatorname{KL}(q\Vert p_v)
+\gamma\operatorname{KL}(q\Vert p_s)\right].
\label{eq:poe_variational}
\end{equation}
Thus, the normalized weighted product yields the distribution closest to
the two experts under this objective.

\par
\endgroup

% Queue the experiment tables in order for the next two page tops.
\begin{table*}[t]
\centering\footnotesize
\setlength{\tabcolsep}{2.0pt}
\renewcommand{\arraystretch}{0.96}
\caption{Average incremental accuracy ($\Aavg$) and final-stage accuracy ($\AT$) under both protocols (\%). Bold marks the best displayed result.}
\label{tab:main_results}
\begin{tabular*}{\textwidth}{@{}l@{\hspace{6pt}}c@{\hspace{8pt}\extracolsep{\fill}}rrrrrrrrrrrr@{\hspace{6pt}}}
\toprule
\multirow{3}{*}{Method} & \multirow{3}{*}{Year} & \multicolumn{4}{c}{ImageNet-R} & \multicolumn{4}{c}{CIFAR-100} & \multicolumn{4}{c}{CUB-200} \\
 & & \multicolumn{2}{c}{B0/Inc20} & \multicolumn{2}{c}{B100/Inc20} & \multicolumn{2}{c}{B0/Inc10} & \multicolumn{2}{c}{B50/Inc10} & \multicolumn{2}{c}{B0/Inc20} & \multicolumn{2}{c}{B100/Inc20} \\
 & & $\Aavg$ & $\AT$ & $\Aavg$ & $\AT$ & $\Aavg$ & $\AT$ & $\Aavg$ & $\AT$ & $\Aavg$ & $\AT$ & $\Aavg$ & $\AT$ \\
\midrule
ZS-CLIP~\cite{radford2021clip} & 2021 & 83.50 & 77.32 & 79.72 & 77.32 & 81.81 & 71.38 & 76.49 & 71.38 & 75.47 & 63.72 & 69.06 & 63.72 \\
SRDA-CLIP~\cite{khawand2023srda} & 2023 & 72.74 & 64.47 & 69.73 & 67.02 & 87.41 & 81.34 & 85.12 & 82.36 & 79.24 & 72.42 & 81.07 & 77.55 \\
FeCAM-CLIP~\cite{goswami2023fecam} & 2023 & 82.72 & 76.43 & 78.73 & 76.43 & 88.16 & 82.00 & 85.00 & 82.00 & 83.18 & 75.23 & 78.81 & 75.23 \\
RAPF~\cite{huang2024rapf} & 2024 & 84.56 & 79.78 & 81.53 & 79.18 & 87.52 & 80.89 & 84.59 & 81.04 & 80.66 & 69.71 & 76.99 & 70.09 \\
SimpleCIL~\cite{zhou2025revisiting} & 2025 & 81.13 & 74.55 & 76.92 & 74.55 & 84.15 & 76.63 & 80.20 & 76.63 & 84.46 & 77.68 & 80.69 & 77.68 \\
ENGINE~\cite{zhou2025engine} & 2025 & 86.23 & 80.45 & 83.69 & 81.08 & 86.89 & 79.29 & 82.82 & 79.23 & 87.32 & 80.05 & 84.01 & 80.43 \\
BOFA~\cite{li2026bofa} & 2026 & 84.82 & 79.17 & 81.59 & 79.55 & 86.37 & 79.51 & 83.61 & 80.08 & 87.41 & 81.53 & 84.23 & 81.50 \\
HASTEN~\cite{hu2026hasten} & 2026 & 85.88 & 80.83 & 82.70 & 80.42 & 88.59 & 82.44 & 85.38 & 82.56 & 86.45 & 80.81 & 83.14 & 80.36 \\
\method & --- & \textbf{88.06} & \textbf{83.05} & \textbf{85.21} & \textbf{83.27} & \textbf{89.84} & \textbf{84.15} & \textbf{86.86} & \textbf{84.19} & \textbf{88.36} & \textbf{82.76} & \textbf{85.16} & \textbf{82.53} \\
\midrule
\multirow{3}{*}{Method} & \multirow{3}{*}{Year} & \multicolumn{4}{c}{Aircraft} & \multicolumn{4}{c}{Cars} & \multicolumn{4}{c}{Food-101} \\
 & & \multicolumn{2}{c}{B0/Inc10} & \multicolumn{2}{c}{B50/Inc10} & \multicolumn{2}{c}{B0/Inc10} & \multicolumn{2}{c}{B50/Inc10} & \multicolumn{2}{c}{B0/Inc10} & \multicolumn{2}{c}{B50/Inc10} \\
 & & $\Aavg$ & $\AT$ & $\Aavg$ & $\AT$ & $\Aavg$ & $\AT$ & $\Aavg$ & $\AT$ & $\Aavg$ & $\AT$ & $\Aavg$ & $\AT$ \\
\midrule
ZS-CLIP~\cite{radford2021clip} & 2021 & 27.06 & 18.48 & 22.44 & 18.48 & 83.89 & 78.51 & 80.39 & 78.51 & 90.20 & 85.46 & 87.75 & 85.46 \\
SRDA-CLIP~\cite{khawand2023srda} & 2023 & 63.57 & 56.56 & 62.88 & 60.22 & 85.06 & 80.01 & 89.88 & 87.79 & 91.45 & 87.57 & 89.94 & 88.10 \\
FeCAM-CLIP~\cite{goswami2023fecam} & 2023 & 65.53 & 57.16 & 59.74 & 57.16 & 92.01 & 87.32 & 89.12 & 87.32 & 92.26 & 88.58 & 90.43 & 88.58 \\
RAPF~\cite{huang2024rapf} & 2024 & 44.32 & 24.06 & 38.68 & 25.56 & 81.86 & 69.28 & 77.41 & 66.52 & 92.30 & 88.04 & 90.48 & 88.10 \\
SimpleCIL~\cite{zhou2025revisiting} & 2025 & 58.48 & 47.94 & 52.17 & 47.94 & 92.29 & 87.35 & 89.35 & 87.35 & 90.31 & 85.50 & 87.91 & 85.50 \\
ENGINE~\cite{zhou2025engine} & 2025 & 69.38 & 58.66 & 64.00 & 58.90 & 94.41 & 90.74 & 92.20 & 90.69 & 92.13 & 87.76 & 89.88 & 87.60 \\
BOFA~\cite{li2026bofa} & 2026 & 70.92 & 60.76 & 67.20 & 62.41 & 94.60 & 90.77 & 92.54 & 91.16 & 91.20 & 86.61 & 89.35 & 87.04 \\
HASTEN~\cite{hu2026hasten} & 2026 & 66.94 & 57.25 & 61.30 & 57.07 & 93.18 & 88.87 & 90.33 & 88.65 & 92.92 & 89.06 & 91.05 & 89.13 \\
\method & --- & \textbf{73.24} & \textbf{64.63} & \textbf{67.55} & \textbf{64.78} & \textbf{94.78} & \textbf{91.31} & \textbf{92.76} & \textbf{91.33} & \textbf{93.19} & \textbf{89.73} & \textbf{91.46} & \textbf{89.72} \\
\midrule
\multirow{3}{*}{Method} & \multirow{3}{*}{Year} & \multicolumn{4}{c}{ObjectNet} & \multicolumn{4}{c}{SUN-397} & \multicolumn{4}{c}{UCF-101} \\
 & & \multicolumn{2}{c}{B0/Inc20} & \multicolumn{2}{c}{B100/Inc20} & \multicolumn{2}{c}{B0/Inc30} & \multicolumn{2}{c}{B150/Inc30} & \multicolumn{2}{c}{B0/Inc10} & \multicolumn{2}{c}{B50/Inc10} \\
 & & $\Aavg$ & $\AT$ & $\Aavg$ & $\AT$ & $\Aavg$ & $\AT$ & $\Aavg$ & $\AT$ & $\Aavg$ & $\AT$ & $\Aavg$ & $\AT$ \\
\midrule
ZS-CLIP~\cite{radford2021clip} & 2021 & 38.43 & 26.43 & 31.12 & 26.43 & 79.45 & 72.14 & 74.98 & 72.14 & 75.88 & 67.79 & 71.68 & 67.79 \\
SRDA-CLIP~\cite{khawand2023srda} & 2023 & 51.96 & 40.63 & 45.93 & 41.19 & 84.81 & 78.94 & 81.52 & 79.00 & 98.00 & 97.01 & 98.81 & 98.14 \\
FeCAM-CLIP~\cite{goswami2023fecam} & 2023 & 51.30 & 39.17 & 44.29 & 39.17 & 84.79 & 79.15 & 81.67 & 79.15 & 98.89 & 98.07 & 98.60 & 98.07 \\
RAPF~\cite{huang2024rapf} & 2024 & 53.93 & 36.01 & 45.79 & 35.68 & 85.06 & 77.91 & 81.41 & 78.13 & 90.42 & 80.90 & 88.06 & 79.54 \\
SimpleCIL~\cite{zhou2025revisiting} & 2025 & 52.06 & 40.13 & 45.11 & 40.10 & 82.15 & 75.60 & 78.64 & 75.60 & 90.41 & 85.79 & 88.08 & 85.79 \\
ENGINE~\cite{zhou2025engine} & 2025 & 59.13 & 45.16 & 51.19 & 44.84 & 85.01 & 78.50 & 81.60 & 78.46 & 94.37 & 89.88 & 92.89 & 90.22 \\
BOFA~\cite{li2026bofa} & 2026 & 59.56 & 47.16 & 52.41 & 47.28 & 85.05 & 78.10 & 82.17 & 79.03 & 94.03 & 89.47 & 93.76 & 90.56 \\
HASTEN~\cite{hu2026hasten} & 2026 & 58.42 & 46.02 & 51.98 & 46.62 & 86.23 & 80.41 & 83.04 & 80.14 & 96.29 & 92.95 & 95.70 & 93.41 \\
\method & --- & \textbf{61.16} & \textbf{49.16} & \textbf{54.36} & \textbf{48.90} & \textbf{87.08} & \textbf{81.67} & \textbf{84.19} & \textbf{81.81} & \textbf{99.03} & \textbf{98.52} & \textbf{98.92} & \textbf{98.64} \\
\bottomrule
\end{tabular*}
\end{table*}

\begin{table*}[t]
  \begin{minipage}[t]{\columnwidth}
    \vspace{0pt}
    \centering\small
\renewcommand{\arraystretch}{1.05}
\setlength{\tabcolsep}{3pt}
\parbox[t][3.3\baselineskip][t]{\linewidth}{%
\caption{Comparison with replay-based methods under B0. Entries are $\Aavg/\AT$ (\%). Bold marks the best displayed result.}
\label{tab:replay_comparison}}
\begin{tabular}{@{}lc@{\hspace{12pt}}c@{\hspace{12pt}}c@{}}
\toprule
\multirow{2}{*}{Method} & \multirow{2}{*}{Images/class} & Aircraft & SUN-397 \\
 & & $\Aavg/\AT$ & $\Aavg/\AT$ \\
\midrule
iCaRL~\cite{rebuffi2017icarl} & 20 & 67.32/57.07 & 63.29/51.50 \\
MEMO~\cite{zhou2023memo} & 20 & 69.34/63.43 & 80.27/72.19 \\
PROOF~\cite{zhou2025proof} & 20 & 63.67/56.20 & 84.05/77.09 \\
\midrule
\method & 0 & \textbf{73.24/64.63} & \textbf{87.08/81.67} \\
\bottomrule
\end{tabular}

  \end{minipage}\hfill
  \begin{minipage}[t]{\columnwidth}
    \vspace{0pt}
    \centering\small
\setlength{\tabcolsep}{2pt}
\renewcommand{\arraystretch}{1.05}
\parbox[t][3.3\baselineskip][t]{\linewidth}{%
\caption{Cumulative component ablations on ImageNet-R B0/Inc20 (\%). Best results are bold.}
\label{tab:component_ablations}}
\begin{tabular}{@{}ccccrr@{}}
\toprule
Per-class & Shared & Posterior & Semantic & \multirow{2}{*}{$\Aavg$} & \multirow{2}{*}{$\AT$} \\
covariance & covariance & fusion & calibration & & \\
\midrule
$\checkmark$ & --- & --- & --- & 78.656 & 71.15 \\
$\checkmark$ & $\checkmark$ & --- & --- & 86.577 & 81.27 \\
$\checkmark$ & $\checkmark$ & $\checkmark$ & --- & 87.920 & 82.88 \\
\midrule
$\checkmark$ & $\checkmark$ & $\checkmark$ & $\checkmark$ & \textbf{88.064} & \textbf{83.05} \\
\bottomrule
\end{tabular}

  \end{minipage}
\end{table*}

\section{Experiments}
\label{sec:experiments}
\subsection{Experimental Setup}
We evaluated our method on nine datasets~\cite{hendrycks2021imagenetr,krizhevsky2009cifar,wah2011cub,maji2013aircraft,krause2013cars,bossard2014food,barbu2019objectnet,xiao2010sun,soomro2012ucf} under two class-incremental
learning protocols, using the class subsets and task splits adopted in
prior work~\cite{zhou2025engine,li2026bofa}.
B0/Inc divides all classes into ten equal-sized tasks, while B50/B100/B150
places half the classes in the initial task and divides the remaining
classes into five equal-sized incremental tasks. The datasets and their
task splits are shown in Table~\ref{tab:main_results}.
Let $A_t$ denote top-1 accuracy over all seen classes after stage $t$.
We report average incremental accuracy
$\Aavg=T^{-1}\sum_{t=1}^{T}A_t$ and final-stage accuracy $\AT$.
Here, $T$ is the total number of tasks, including the initial task.

Our implementation uses CLIP ViT-B/16 with both the image and text encoders frozen.
Input images had a resolution of $224\times224$, and the extracted
512-dimensional image features were $\ell_2$-normalized. At each incremental
task, we used current-task training images to compute visual statistics
and learned new-class text-template weights with Adam using a batch size
of 128. Template weights were initialized uniformly, while old-class text
representations remained unchanged. No old images were replayed during training.
The class-order random seed was fixed to 1993.

The compared baselines fall into three groups: (i) direct frozen-feature classifiers,
ZS-CLIP and SimpleCIL, based on image-text matching and class-mean prototypes,
respectively; (ii) covariance-based statistical classifiers, SRDA-CLIP and
FeCAM-CLIP, adapted to frozen CLIP image features; and (iii) CLIP-based
incremental adaptation methods, RAPF, ENGINE, BOFA, and HASTEN.
All baseline results come from complete local CLIP-based runs with the same
class subsets, task splits, and class order.

% Queue the analysis figures before their discussion to keep them in the main text.
\begin{figure*}[t]
  \captionsetup{skip=4pt}
  \begin{minipage}[t]{\columnwidth}
    \vspace{0pt}
      \centering
  \parbox[t][0.72\linewidth][t]{\linewidth}{%
    \includegraphics[width=\linewidth,height=0.72\linewidth,keepaspectratio]{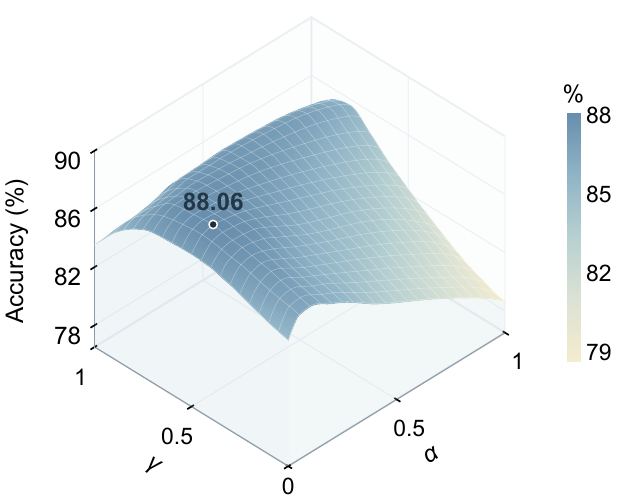}}
  \captionof{figure}{Hyperparameter sensitivity on ImageNet-R B0. The surface interpolates 120 measured configurations; the dot marks the reported setting.}
  \label{fig:sensitivity_imagenetr}

  \end{minipage}\hfill
  \begin{minipage}[t]{\columnwidth}
    \vspace{0pt}
      \centering
  \parbox[t][0.72\linewidth][t]{\linewidth}{%
    \includegraphics[width=\linewidth]{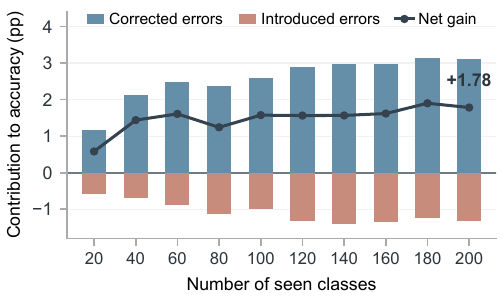}}
  \captionof{figure}{Accuracy gains and losses from posterior fusion on ImageNet-R B0. Bars show corrected and introduced errors; the curve shows the net gain.}
  \label{fig:complementarity_imagenetr}

  \end{minipage}
\end{figure*}

\subsection{Comparison with State-of-the-Art Methods}
With both CLIP encoders frozen, \method achieved state-of-the-art performance
among the compared methods under both protocols, obtaining the highest
average incremental and final-stage accuracies averaged over the nine datasets
(Table~\ref{tab:main_results}). Both metrics were also highest in each of
the 18 dataset-protocol settings. Under B0, these metrics reached 86.08\%
and 80.55\%, surpassing the strongest baseline in the table for each metric by
2.21 and 2.93 percentage points, respectively. When the initial task
contained half the classes, they reached 82.94\% and 80.57\%, exceeding
the strongest baseline by 2.18 and 2.95 percentage points, respectively.
\method substantially outperformed the covariance-based statistical
classifiers SRDA-CLIP and FeCAM-CLIP on both metrics averaged across the
nine datasets, while also achieving higher scores on both metrics in every
dataset.
Under both protocols, the cross-dataset gain in final-stage accuracy exceeded
that in average incremental accuracy, indicating that the advantage remained
after all classes had been introduced.

We further compare \method with the replay-based methods iCaRL, MEMO, and PROOF,
which retained 20 training images per class, on Aircraft B0/Inc10 and
SUN-397 B0/Inc30 (Table~\ref{tab:replay_comparison}). All baseline results
are obtained from complete local CLIP-based runs using the same data splits and
class order as our method. Compared with MEMO on Aircraft and PROOF on
SUN-397, the strongest local replay baselines on the respective datasets,
\method improved average incremental accuracy by 3.90 and 3.04 percentage
points and final-stage accuracy by 1.20 and 4.58 points, respectively.
These results show that, without retaining or replaying old images, our method
can outperform the compared replay-based methods even when these baselines retain
20 training images per class.

\subsection{Further Analysis}
We analyzed component contributions, hyperparameter sensitivity,
prediction complementarity, and computational efficiency.

\paragraph{Component Ablation.}
We conducted cumulative ablations on ImageNet-R under B0/Inc20, starting with a visual expert
using independently estimated class-specific covariances ($\alpha=1$) and
successively adding shared covariance, visual-semantic posterior fusion,
and class-adaptive semantic calibration while holding all other settings fixed.
Table~\ref{tab:component_ablations} reports average incremental and final-stage
accuracies. Shared covariance increased average incremental accuracy by 7.921 percentage
points, providing the largest gain and supporting the use of shared
statistics to stabilize class-specific covariance estimation.
Visual-semantic posterior fusion then improved average incremental accuracy
by a further 1.343 points, indicating that semantic predictions complement
the evidence provided by visual distributions. Finally, class-adaptive
semantic calibration added another 0.144 points by learning class-specific
template weights from current-task images, demonstrating a small additional
benefit from adapting class text representations to downstream images.

\paragraph{Hyperparameter Analysis.}
We analyzed the effects of the covariance mixing coefficient $\alpha$ and
fusion coefficient $\gamma$ on ImageNet-R under B0, keeping all other settings
fixed. As shown in Figure~\ref{fig:sensitivity_imagenetr}, appropriately
combining class-specific and shared covariances, together with visual and
semantic predictions, yields high average incremental accuracy. Specifically,
with only the visual expert ($\gamma=0$), changing $\alpha$ from 1 to 0.1
increased accuracy from 78.66\% to 86.58\%, which also exceeds the 85.19\%
obtained with shared covariance alone. This indicates that both stable
estimation and retaining class-specific variation benefit classification.
With $\alpha=0.1$ fixed, incorporating semantic predictions at $\gamma=0.5$
further increased accuracy to 88.06\%, whereas relying entirely on the
semantic expert ($\gamma=1$) yielded 83.54\%, demonstrating the complementarity
between the two experts. Moreover, with $\gamma=0.5$ fixed, the neighboring values
$\alpha=0.05$ and $0.2$ achieved 87.86\% and 87.85\%, respectively;
with $\alpha=0.1$ fixed, $\gamma=0.4$ and $0.6$ achieved 87.83\% and
87.97\%, respectively. These results indicate that high accuracy is not
confined to a single configuration. This suggests local
robustness to parameter choices within the evaluated neighborhood.

\paragraph{Complementarity of Visual and Semantic Predictions.}
To evaluate the effectiveness of visual-semantic posterior fusion, we
analyzed per-sample prediction changes before and after fusion on ImageNet-R.
As shown in Figure~\ref{fig:complementarity_imagenetr}, posterior fusion
provided positive accuracy gains at every incremental stage. At the final
stage, fusion changed incorrect visual predictions to correct ones for
3.10\% of the test samples. After accounting for the reverse change on
1.32\% of the samples, the net accuracy gain was 1.78 percentage points.
On average over stages, the visual expert, semantic expert, and their
fusion achieved accuracies of 86.58\%, 83.54\%, and 88.06\%, respectively.
At the final stage, fusion also correctly classified 30 images that neither
expert recognized with its own top-1 prediction. By combining the full
class scores, fusion can exploit information not reflected in either
expert's top-1 label.
These results show that semantic predictions provide complementary
discriminative information for samples that the visual expert struggles
to classify correctly, enabling posterior fusion to consistently improve
classification performance as the number of seen classes increases.

\paragraph{Computational Efficiency.}
\begin{table}[!htbp]
\centering\small
\renewcommand{\arraystretch}{1.12}
\setlength{\tabcolsep}{3pt}
\caption{Accuracy and efficiency on ImageNet-R B0 using the same RTX 3090. Train: total training time; Eval: time per image including feature extraction; GPU: peak allocated memory.}
\label{tab:efficiency}
\begin{tabular}{@{}l@{\hspace{12pt}}c@{\hspace{12pt}}c@{\hspace{12pt}}c@{\hspace{12pt}}c@{}}
\toprule
\multirow{2}{*}{Method} & $\Aavg/\AT\uparrow$ & Train$\downarrow$ & Eval$\downarrow$ & GPU$\downarrow$ \\
 & (\%) & (s) & (ms) & (MiB) \\
\midrule
ENGINE & 86.23/80.45 & 2410.2 & 39.37 & 5312.9 \\
BOFA & 84.82/79.17 & 1728.5 & 5.92 & 3175.3 \\
\midrule
\method & \textbf{88.06/83.05} & \textbf{161.3} & \textbf{3.15} & \textbf{2181.5} \\
\bottomrule
\end{tabular}
\end{table}

During training, as shown in Table~\ref{tab:efficiency}, completing all ten
ImageNet-R stages on the same RTX 3090 took 161.3\,s for \method,
compared with 2410.2\,s for ENGINE and 1728.5\,s for BOFA.
Thus, \method required approximately 6.7\% and 9.3\% of their
training time, respectively.
Its peak allocated GPU memory was 58.9\% and 31.3\% lower, respectively,
demonstrating very low training overhead relative to these two methods. Our lightweight optimization keeps both CLIP
encoders frozen and optimizes only a single $K$-dimensional vector of template
weight parameters per new class, without introducing task-specific adapters
or learnable prompts. Thus, gradient optimization involves only a small
number of parameters and avoids backpropagation through the encoders and
the associated activation-storage cost.

During evaluation, visual scoring reuses precomputed precision matrices
and log-determinants, while posterior fusion directly combines the two
experts' outputs. Including feature extraction, \method required
3.15\,ms per test-image visit, compared with 39.37\,ms for ENGINE and
5.92\,ms for BOFA, reducing per-image evaluation time by 92.0\% and
46.8\%, respectively. These results show that class-distribution modeling
and two-expert fusion can maintain efficient inference. Both experts share
a single image-encoding pass and use cached visual scoring parameters and
class text representations. Fusion then operates directly on their class
scores, without additional encoder passes or an iterative optimization procedure.

\section{Conclusion}
\label{sec:conclusion}

We present \method, a framework with lightweight optimization for
exemplar-free class-incremental learning that combines class-conditional
visual distribution modeling, class-adaptive semantic calibration, and
visual-semantic posterior fusion with frozen CLIP encoders.
Across nine datasets under two incremental protocols, it achieves
state-of-the-art performance, attaining the highest average incremental
and final-stage accuracies among the compared methods when averaged across
the nine datasets.
These results show that fully exploiting visual and semantic information
in pre-trained representations enables effective incremental recognition
without continual encoder adaptation.

\section*{LLM Usage Statement}
OpenAI Codex assisted with language polishing, experimental planning,
code development for experiments, and result analysis. The authors are
responsible for the final manuscript, implementation, and reported results.

{
    \small
    \raggedbottom
    \setlength{\bibsep}{1pt}

}

\end{document}